\documentclass[runningheads]{llncs}
\usepackage{graphicx}
\usepackage{booktabs,multirow}
\usepackage{amssymb}
\usepackage{amsmath}
\usepackage{amstext}
\usepackage{comment}
\usepackage{color}
\usepackage{pifont}
\usepackage{bbding}
\usepackage{amsfonts,amssymb,amsmath}

\usepackage[pagebackref=true,breaklinks=true,colorlinks=true,bookmarks=false,citecolor=blue,urlcolor=black,linkcolor=blue]{hyperref}

\usepackage{todonotes}

\usepackage{amsmath}

\usepackage[marginal]{footmisc}

\begin{document}
\title{E-DSSR: Efficient Dynamic Surgical Scene Reconstruction with Transformer-based \\Stereoscopic Depth Perception}
%
%

\titlerunning{E-DSSR: Efficient Dynamic Surgical Scene Reconstruction}


\author{Yonghao Long \inst{*1}\Envelope \and Zhaoshuo Li\inst{*2} \and Chi Hang Yee\inst{3} \and \\ Chi Fai Ng \inst{3} \and Russell H. Taylor\inst{2} \and  Mathias Unberath\inst{2} \and Qi Dou\inst{1,4}
}
\authorrunning{Y. Long et al.}
\institute{Dept. of Computer Science and Engineering, The Chinese University of Hong Kong\\
\email{yhlong@cse.cuhk.edu.hk} \\
\and Department of Computer Science, Johns Hopkins University
\and SH Ho Urology Centre, Dept. of Surgery, The Chinese University of Hong Kong
\and T Stone Robotics Institute, The Chinese University of Hong Kong
}
\maketitle              
\begin{abstract}

Reconstructing the scene of robotic surgery from the stereo endoscopic video is an important and promising topic in surgical data science, which potentially supports many applications such as surgical visual perception, robotic surgery education and intra-operative context awareness. However, current methods are mostly restricted to reconstructing static anatomy assuming no tissue deformation, tool occlusion and de-occlusion, and camera movement. However, these assumptions are not always satisfied in minimal invasive robotic surgeries. In this work, we present an efficient reconstruction pipeline for highly dynamic surgical scenes that runs at 28 fps. Specifically, we design a transformer-based stereoscopic depth perception for efficient depth estimation and a light-weight tool segmentor to handle tool occlusion. After that, a dynamic reconstruction algorithm which can estimate the tissue deformation and camera movement, and aggregate the information over time is proposed for surgical scene reconstruction. We evaluate the proposed pipeline on two datasets, the public Hamlyn Centre Endoscopic Video Dataset and our in-house DaVinci robotic surgery dataset. The results demonstrate that our method can recover the scene obstructed by the surgical tool and handle the movement of camera in realistic surgical scenarios effectively at real-time speed.

\end{abstract}

\section{Introduction}
\begin{figure}[t]
	\centering
	\includegraphics[width=0.9\linewidth]{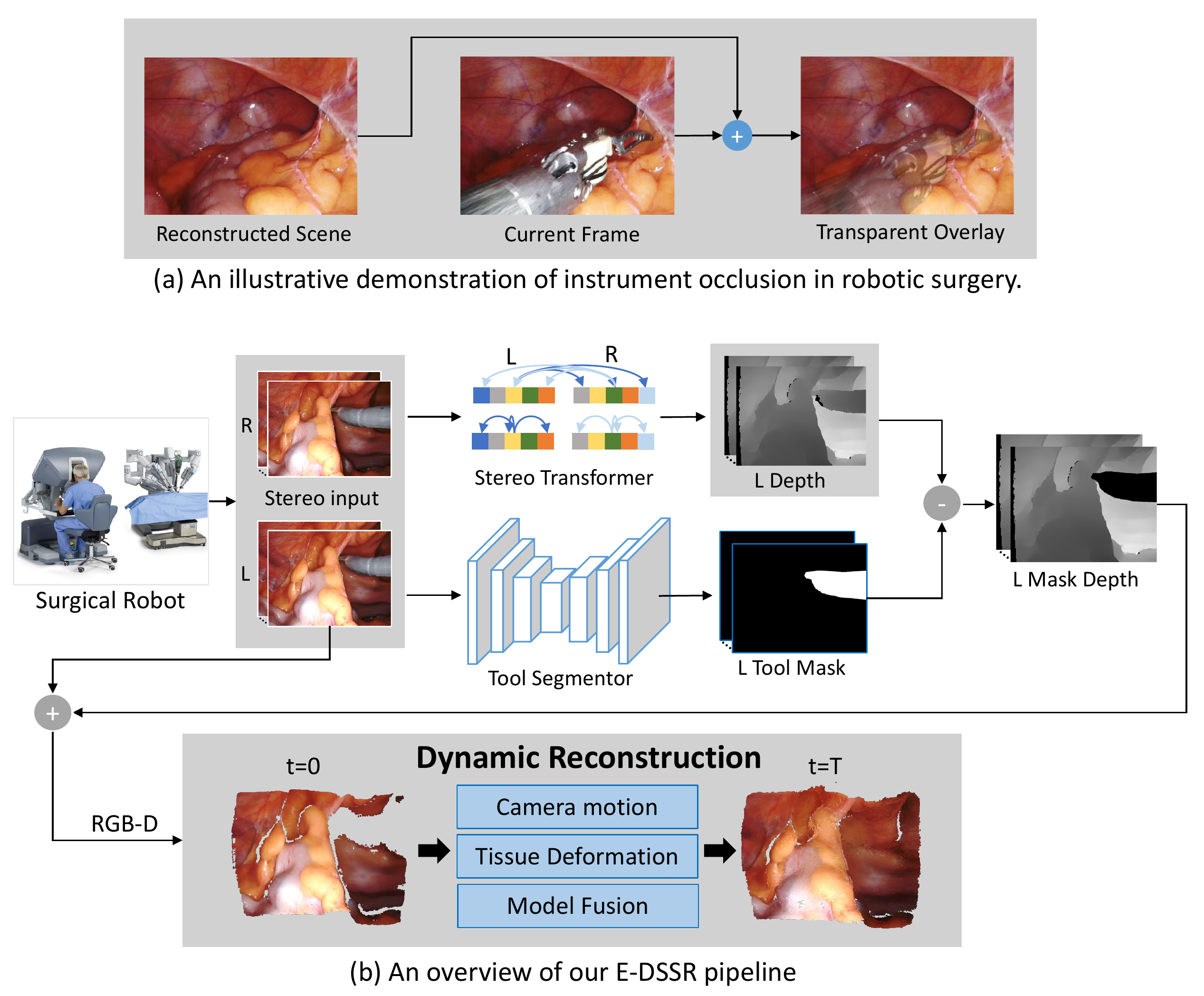}
	\caption{Illustration of the insight (a) and the pipeline of our proposed E-DSSR (b).}
	\label{fig:framework}
\end{figure}

Reconstructing the surgical scene from stereo endoscopic video in robotic-assisted minimally invasive surgeries (MIS) is an important topic as it is central to downstream tasks. 
\footnote{\noindent *Authors contributed equally to this work.}
For example, during surgical training, it is desirable to expose the trainees to the complete soft-tissue even if surgical tools  block the view partially~\cite{stoyanov2008intra,taylor2016medical} in order to provide enriched context for understanding the surgical manipulation. As illustrated in \autoref{fig:framework}(a), given the reconstruction from recorded surgical videos and the current video frame with instrument blocking the view, a transparent overlay can be generated for AI-augmented demonstration which brings new possibilities for robotic surgery education. A similar method potentially may be used to provide additional useful context intraoperatively.

However, reconstruction of the surgical scene in laparoscopy is challenging for three reasons. First, the soft-tissue is constantly deforming. This gives rise to challenges to soft-tissue localization. Secondly, it presents heavy occlusion and dynamic movement of the surgical instruments. Identification of occlusion and proper handling of de-occlusion require spatial and temporal coherence and consistency. Lastly, the changes of camera view points compound the aforementioned difficulties into an ego motion task with dynamic objects. Even though there exists prior works in 3D reconstruction in surgical scene, they are generally limited by assuming a static scene~\cite{liu2020reconstructing} or no presence of surgical tools~\cite{song20203d}. 

Closest to our work are~\cite{li2020super,lu2020super}, which jointly handle tissue deformation and surgical tool occlusion but using kinematics information. We improve upon prior work with an \textit{image-only} reconstruction pipeline shown in \autoref{fig:framework}(b) that improves the quality and run-time speed. We implement an efficient transformer-based depth perception module and a light-weight tool segmentor to reconstruct the surgical scenes with only stereo endoscopic image frames as inputs. The two modules run in parallel to output a masked depth estimation without surgical instruments. The masked depth map is later used to produce a temporally and spatially coherent reconstruction of the soft-tissue. We also demonstrate the effectiveness of our pipeline with camera motion and smoke, which is missing from~\cite{li2020super,lu2020super}.

Our main contributions are summarized as follows: We propose a novel online reconstruction pipeline called E-DSSR which can reconstruct the surgical scene with \textit{only} stereo endoscopic videos as input, and handle the cases of tissue deformation, tool occlusion and the camera movement simultaneously. We qualitatively and quantitatively evaluate our proposed pipeline on both the public Hamlyn Centre Endoscopic Video Dataset (Hamlyn Dataset)~\cite{ye2017self} and our in-house DaVinci robotic surgery data to validate the effectiveness of our method.

\section{Related Work}
Recent works of~\cite{lamarca2020defslam,song2018mis} have explored SLAM system to handle the non-rigidity of tissues. However, they assume a simplified environment where surgical tools are not present. Later, Li et al. \cite{li2020super} proposes a reconstruction framework to simultaneously reconstruct the soft tissue and track surgical instruments using kinematics. However, instrument location is prone to noise and error given the long kinematics chain as demonstrated in \cite{ferguson2020comparing}. Lu et al. \cite{lu2020super} further improves the previous framework with deep learning methods to estimate depth, and a key-point+kinematics hybrid approach to localize the surgical instrument. However, the result is only demonstrated in an \textit{ex vivo} environment with a fixed camera pose, without camera motion and realistic artifacts such as blood. Furthermore, the surgical instrument is removed by computing the pose of the surgical tool and then rendering a binary mask from a 3D model, which is slower as we will show in \autoref{ssec:quan_eval}. In this work, we present an efficient reconstruction pipeline that only uses a stream of stereo images as input, as well as being capable of handling camera motion and surgical effects such as smoke in addition to the previous challenges.

\section{Method}

\autoref{fig:framework}(b) shows the overview of our proposed dynamic surgical scene reconstruction pipeline E-DSSR. 
We denote left (L) and right (R) RGB image at time stamp $t\in\left[0,T\right]$ as $l_t$ and $r_t$. 
A transformer-based stereo depth estimation estimates the depth image $d_t$ using $l_t$ and $r_t$. A tool segmenting network predicts the mask of the tool $l_t^m$ concurrently. Note that both networks are designed to be light-weight to enable real-time performance. The masked depth image $d_t^m$ and left frame $l_t$ from $t=0$ to $t=T$ are input to a dynamic reconstruction algorithm which can dynamically recover the 3D information of the surgical scene.


\subsection{Light-weight Stereo Depth Estimation and Tool Segmentation}
\label{ssec:stereo}

\subsubsection{Transformer-based depth estimation.} To reconstruct the soft-tissue with high quality, it is important to only include depth estimations with high confidence. In the occluded region where pixels are not observed commonly by both $l_t$ and $r_t$, depth cannot be accurately estimated.  
Therefore, we opt for the recently proposed Stereo Transformer (STTR)~\cite{li2020revisiting} as the depth estimation module as it explicitly identifies these regions. STTR densely compares pixels in $l_t$ and $r_t$ along epipolar lines to find the best matches to estimate depth. STTR uses the attention mechanism~\cite{vaswani2017attention}, which computes the attention (feature similarities $\alpha$) between source $p_{src}$ and target $p_{tgt}$ and outputs the updated features $p_{out}$ as:
\begin{align}
    p_{out} &= \text{softmax}(\frac{\alpha}{\sqrt{C_{attn}}}) W_v p_{tgt} \text{, with } \alpha = p_{src}^T W_q^T W_k p_{tgt}.
\end{align}
where $W_q, W_k, W_v$ are the learnt projection weights of dimension $\mathbb{R}^{C_{attn} \times C_{attn}}$ to transform the inputs to an embedding space. After the final attention, the similarities between the pixels from $l_t$ and $r_t$ are used as likelihood for pixel matching. 
We propose a new variant of STTR, to enable faster computational speed while without harming performance. We note that a large amount of FLOPS of STTR is within the attention module. Since STTR is comparing the pixels along epipolar lines (of $W^2$ potential matches) for $H$ lines, its number of parameters and FLOPS are in the order of
\begin{align}
    \text{Number of parameters} = \mathcal{O}(C_{attn}^2 N_{attn}),  \text{ FLOPS} = \mathcal{O}(H W^2 N_{attn}).
\end{align}
where $H, W$ are the height and width of the input image, and $N_{attn}$ is the number of attentions computed. To avoid significant deterioration of the performance, we keep number of parameters by making $C_{attn}N_{attn}$ constant while quartering the number of attentions. Following prior work, we train STTR on the large-scale synthetic Scene Flow dataset~\cite{mayer2016large}, which is a commonly used pre-training dataset for stereo networks. 
We show that transformer-based depth module improves the reconstruction quality in surgical scene quantitatively in \autoref{ssec:quan_eval} and our light-weight STTR is faster without much performance sacrifice. This is also the first time transformer-based module is applied to surgical scene depth estimation.

\subsubsection{Efficient surgical tool segmentation.}
Given depth estimation of the scene, it is desirable to isolate the depth of soft-tissue from surgical tools since surgical tools are considered as ``outlier'' in soft-tissue reconstruction.
For this purpose, we predict a binary mask indicating which pixels belong to the tools. 
U-Net~\cite{ronneberger2015u} has been proven to be a light-weight yet accurate model for segmentation tasks
~\cite{allan20192017,jin2019incorporating}. In our pipeline, we design a light-weight U-Net with VGG11~\cite{simonyan2014very} as backbone and with 5 scales of down-sample layers to maintain a run-time of around 12ms per frame (input resolution of 640$\times$512). 
We train the model on public dataset of Robotic Instrument Segmentation from the 2017 MICCAI EndoVis Challenge~\cite{allan20192017} and directly use it to predict binary instrument masks on our robotic surgery datasets.
To mitigate the performance variation of the trained U-Net due to slight domain gap between the training data and the data we evaluate our pipeline on, we perform morphological operation on the segmentation mask to refine tool segmentation boundaries. 

\subsection{Dynamic Reconstruction}

\subsubsection{Surfel representation.} Different from most existing methods that adopt a volumetric model~\cite{newcombe2015dynamicfusion}, we rely on a memory-efficient data representation surfel~\cite{gao2019surfelwarp}, which is suitable to the varying environment in surgical scenes. Surfels are tuples of variables including 3D position $v$, surface normal $n$, confidence $c$ and timestamp of last observation $t$ stored as an unordered list $S$. In our pipeline, position $v$ is computed using inverse camera intrinsic matrix and estimated masked depth $d_t^m$. Confidence $c$ is computed using radial distance from the camera center, with the intuition that more oblique points will exhibit larger uncertainty. The canonical surfels $S_{ref}$ is stored in the coordinate defined by the first frame observed, which is then continuously updated by incorporating the surfels $S_{obs}$ of newly observed frames.

\subsubsection{Camera pose.} The stereoscope will move during the surgery and the movement would not be continuous nor fast to avoid harming tissue based on consultation with clinical collaborators. As a result, we can assume that the motion between adjacent frames is relatively small, $S_{ref}$ is transformed to the current view given the most recent camera pose and projected to the camera plane. Between $S_{ref}$ and $S_{obs}$, if the surfels' normals and depth values $d$ are closer than a threshold, a correspondence is found. Given all correspondences $P$, the camera pose $T_{cam}$ is solved by minimizing the energy function defined as follows:
\begin{align}
    E_{depth}(T_{cam}) = \sum_{(S_{ref}, S_{obs}) \in P} (n_{obs}^T(d_{ref}-d_{obs}))^2.
\label{eqn:depth}
\end{align}

\subsubsection{Tissue deformation field.} As the surgical scene and tissue is non-rigid and dynamic, to efficiently model the deformation, a sparse node graph $\mathcal{W}$~\cite{li2009robust} is built with each node represented by its position $v$ similar to surfels, but also a local 6 DoF deformation $T$. Given this set of sparse nodes, any points can interpolate its deformation using a weighted sum of node deformation based on the distance and radius. For a given query position $x$ and node $j$ in $\mathcal{W}$, the weight can be found as $w_j(x)=\text{exp}(-\frac{||v_j-x||^2}{2r_j^2})$. After applying the updated rigid motion of the camera from previous section, the deformations in node graph $\mathcal{W}$ is solved using following equation with as-rigid-as-possible regularization:
\begin{align}
    E_{deform}(\mathcal{W}) = E_{depth}(\mathcal{W}) + \sum_{j\in\mathcal{W}}\sum_{i\in N_j}||T_jv_j-T_iv_j||^2.
\end{align}
where $N_j$ is the neighboring nodes of node $j$, with the intuition that motion within a neighborhood should be as small as possible.

\subsubsection{Model Fusion.}
Lastly, to recover and reconstruct the whole surgical scene, the observation will get integrated to the canonical model $S_{ref}$. For surfels with correspondences, they are fused as one where confidence values get accumulated and the timestamp gets updated. The normal and position are updated as the confidence-weighted sum of observed points and model points.
For the fused surfel to be added to the canonical model, the sum of confidence in local neighborhood needs to be higher than a pre-defined threshold and local motion needs to be consistent. At the same time, a surfel in the canonical model will be removed if it is not observed for a long time. 
\section{Experiments}
\begin{figure}[t]
	\centering
	\includegraphics[width=0.9\textwidth]{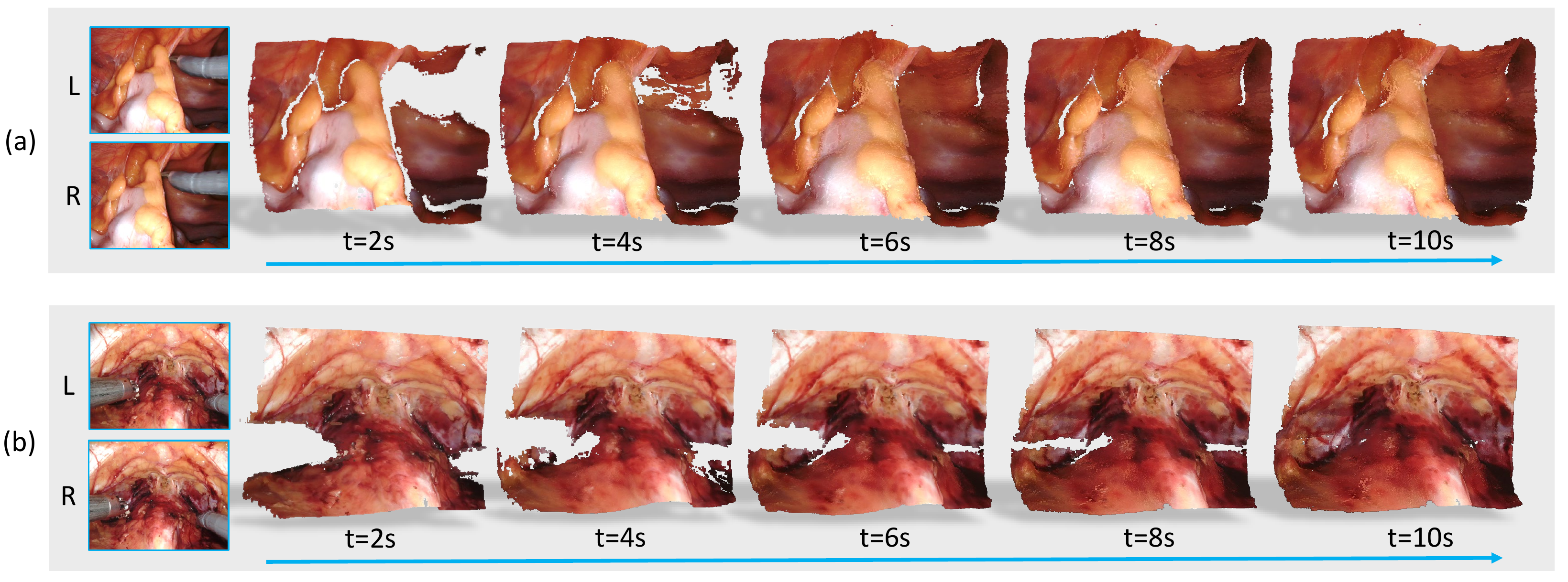}
	\caption{Qualitative result of the in-house DaVinci dataset. The first column shows the stereo RGB frames when $t=0$, while the other columns show the result of the reconstructed result along time axis.}
	\label{fig:inhouse_result}
\end{figure}
\subsection{Experimental Setting}
We evaluate the effectiveness of our proposed dynamic reconstruction framework on two datasets: (1) the public Hamlyn Dataset~\cite{ye2017self}, and (2) our in-house DaVinci robotic surgery dataset of prostatectomy procedure.
The Hamlyn Dataset consists of rectified stereo images with resolution of $384\times192$ collected in partial nephrectomy and without camera calibration information.
Our in-house dataset contains 6 cases of high-resolution stereo videos, each records the whole procedure of robotic prostatectomy. We employ the method of Zhang \textit{et al.}~\cite{zhang2000flexible} to calculate the camera calibration information of our surgical stereoscope.


We collected 5 video clips (with 1200 pairs of rectified stereo frames) from our in-house dataset and 2 video clips (600 pairs) from Hamlyn Dataset. Each clip lasts for around 10 seconds, including scenarios of surgical tool occlusion, camera movement and tissue deformation. All the clips are used for testing, as all components in our method do not require the clips for training. In the experiment, we downsample the video frames of our in-house dataset from the original resolution $1280\times1024$ to $640\times512$. 
We implement the tool segmentation network and the transformer-based depth estimation with PyTorch~\cite{paszke2019pytorch}. The dynamic reconstruction is implemented using C++ with CUDA~\cite{sanders2010cuda} to accelerate the running speed. Our experiment is conduced on a PC with one Nvidia TITAN RTX GPU and Intel Xeon(R) W-2123 CPU (3.60GHz $\times$ 8).

\subsection{Qualitative Result}

To demonstrate the effectiveness of the reconstruction pipeline, we show two examples in \autoref{fig:inhouse_result} from our in-house dataset. In \autoref{fig:inhouse_result}, the scenes are reconstructed with a fixed camera view, moving surgical tools, and tissue deformations. It can be shown from \autoref{fig:inhouse_result} that our pipeline can recover the blocked surgical scene, and continuously improve and complete the canonical model $S_{ref}$ given subsequent frames which expose the obstructed tissue. 

\begin{figure}[t]
	\centering
	\includegraphics[width=0.9\textwidth]{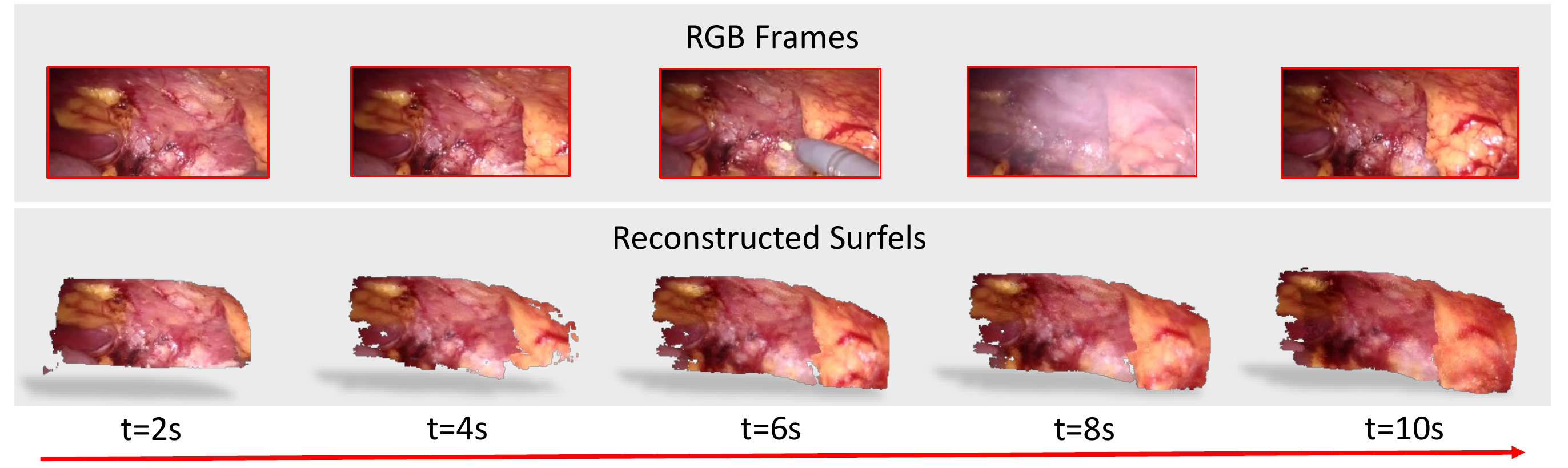}
	\caption{Qualitative results on the Hamlyn Dataset. The first row shows the RGB frames while the second row shows the reconstructed result along time axis.}
	\label{fig:hamlyn_result}
\end{figure}

We show another example in \autoref{fig:hamlyn_result} from the Hamlyn Dataset where the recorded video is more challenging with camera movement, surgical tool movement, tissue deformation and smoke. As shown in \autoref{fig:hamlyn_result}, even with camera movement, the dynamic reconstruction algorithm can still track the soft-tissue and complete the surgical scene of the canonical model $S_{ref}$. Furthermore, the reconstruction pipeline is also robust against smoke. 

\subsection{Quantitative Evaluation and Analysis}
\label{ssec:quan_eval}

Since acquiring the 3D model of the tissue or ground-truth depth with additional sensors in an \textit{in vivo} environment is currently impractical due to clinical regulations, we use image-based metrics to evaluate the dynamic reconstruction results. Structural Similarity Index Measure (SSIM)~\cite{wang2004image} has been widely used for computing consistencies between two images, with a higher value indicating a higher similarity. Peak-Signal-to-Noise Ratio (PSNR)~\cite{hore2010image} is a metric used to estimate the distortion between target image and synthesized (or noisy) image with a higher value indicating a smaller distortion. Both metrics have been used as evaluation metrics to assess the similarities of the predicted (re-projected) image and original image in absence of ground truth depth information~\cite{godard2017unsupervised,li2020unsupervised}. Following prior works, we adopt them to evaluate the performance of our dynamic reconstruction results.

\begin{table}[t]
	\centering
	\caption{The quantitative evaluation of the dynamic reconstruction. Note that the reported time is tested on processing one frame with resolution of $640\times512$. Asterisk sign indicates our proposed method.}
	\label{tab:comparison}
    \resizebox{\textwidth}{!}{
		\begin{tabular}{l|cc|cc|c}
			\toprule[1.5pt]
			\multirow{2}{*}{Method} & \multicolumn{2}{c|}{in-house dataset} & \multicolumn{2}{c|}{Hamlyn Dataset} & \multirow{2}{*}{Speed}
			\\ \cline{2-5} 
			& $~~\text{SSIM}_a(\%)$~~  & ~~$\text{PSNR}_a$~~  & ~~$\text{SSIM}_a(\%)$~~ & ~~$\text{PSNR}_a$~~ \\ \midrule
			HSM~\cite{yang2019hierarchical} + DR w/o mask            
			& 58.48 $\pm$ 6.82   & 11.14 $\pm$ 1.83    & 37.98 $\pm$ 7.32      & 10.27 $\pm$ 1.70   &18Hz \\
			HSM~\cite{yang2019hierarchical} + DR w/ mask          
			& 59.10 $\pm$ 6.71  & 11.76 $\pm$ 1.67    & 38.39 $\pm$ 7.10      & 10.55 $\pm$ 1.86   &15Hz  \\
			E-DSSR w/o mask             
			& 64.17 $\pm$ 4.67   & 13.59 $\pm$ 1.72    & 40.83 $\pm$ 8.45      & 12.01 $\pm$ 2.05   &36Hz \\
			DSSR w/o mask           
			& 65.09 $\pm$ 5.64   & 13.00 $\pm$ 1.61    & 41.83 $\pm$ 7.20      & 13.04 $\pm$ 2.07 &18Hz   \\
            \midrule
			\textbf{E-DSSR* (efficient)}    
			& 66.65 $\pm$ 4.59 & $\mathbf{13.68 \pm 1.65}$ & 41.97 $\pm$ 7.32 & 12.85 $\pm$ 2.03 & 28Hz\\
            \textbf{DSSR* (high-quality)}          
			& $\mathbf{66.81 \pm 4.90}$   & 13.64 $\pm$ 1.81    & $\mathbf{42.41 \pm 7.12}$    & $\mathbf{13.09 \pm 2.14}$  &15Hz \\  \bottomrule[1.5pt]
		\end{tabular}
	}
\end{table}

Specifically, we compare the similarities between the observed frame $l_t$ and the re-projected frame $\hat{l}_t$ using the reconstructed canonical surfel. Since our goal is to reconstruct soft-tissue, the mask of the surgical tool is applied to the predicted frame to obtain the masked frames $l_t^m$ and $\hat{l}_t^m$ with only tissue information. We computed the average SSIM and PSNR across all the frames and for all video clips as the final evaluation metrics, which is shown below:
\begin{small}
\begin{equation}
    \text{SSIM}_a = \frac{1}{T} \sum_{t \in \mathcal{T}}SSIM(l_t^m,\hat{l}_t^m);~~
    \text{PSNR}_a = \frac{1}{T} \sum_{t \in \mathcal{T}}PSNR(l_t^m,\hat{l}_t^m).
\end{equation}
\end{small}
\\
To demonstrate the effectiveness of our method and assess the contribution of each component in our pipeline, we conduct a set of experiments with our proposed E-DSSR pipeline. We compare E-DSSR with a similar pipeline DSSR with our light-weight STTR replaced by STTR. We further compare with a reconstruction pipeline using a different depth estimation network HSM~\cite{yang2019hierarchical}, which is a state-of-the-art fully convolutional stereo depth estimation method. 
The full list of experiments are shown in ~\autoref{tab:comparison}, including: (1) HSM + Dynamic Reconstruction (DR) without tool mask, (2) HSM + Dynamic Reconstruction (DR) with tool masked, (3) DSSR without tool mask, (4) E-DSSR without tool mask, and our proposed methods (5) DSSR and (6) E-DSSR. 

We evaluate the advantages of using transformer-based depth module by comparing E-DSSR with HSM + DR w/ mask, we can see that our method outperforms by 7.55\% $\text{SSIM}_a$ and 1.92 $\text{PSNR}_a$ for the in-house dataset and by 3.58\% $\text{SSIM}_a$ and 2.30 $\text{PSNR}_a$ for Hamlyn Dataset.

We also evaluate the contribution of tool segmentation module in the reconstruction result. By comparing E-DSSR and E-DSSR w/o mask, we show that without explicit tool identification, the $\text{SSIM}_a$ drops from 66.65\% to 64.17\% and $\text{PSNR}_a$ drops from 13.68 to 13.59 for in-house dataset. As for the Hamlyn Dataset, the $\text{SSIM}_a$ drops from 41.97\% to 40.83\% and $\text{PSNR}_a$ drops from 12.85 to 12.01. The same observation can be found by comparing DSSR and DSSR w/o mask, HSM + DR w/ mask and HSM + DR w/o mask. All results demonstrate the benefit of the proposed tool segmentation module. 

Finally, while DSSR achieves the best result of $\text{SSIM}_a$ on both dataset and $\text{PSNR}_a$ on Hamlyn Dataset, the E-DSSR is nearly two times faster than the DSSR (28Hz v.s. 15Hz) and achieves best result of $\text{PSNR}_a$ on in-house dataset while with little performance compromise (0.16\% $\text{SSIM}_a$ on in-house dataset, 0.44\% $\text{SSIM}_a$ and 0.24 $\text{PSNR}_a$ on Hamlyn Dataset).

We also compare our approach with the previous proposed work~\cite{li2020super}. Based on their reported result, our light-weight pipeline is 14$\times$ faster (28Hz v.s. 2Hz) given the same image resolution. Comparing our image-based tool segmentation module with their kinematics-driven 3D model rendering approach, our module is more than 5 times faster (83Hz v.s. 15Hz). 


\section{Conclusion}
In conclusion, we propose an efficient, image-only reconstruction pipeline for surgical scenes. Our light-weight modules enable real-time reconstruction result in the presence of camera motion, surgical tools, and tissue deformation. We evaluate our pipeline qualitatively and quantitatively to demonstrate its effectiveness. We did not evaluate on longer duration, because the camera views may be completely different within a long video clip, which could be considered as several short clips to be reconstructed. 
Future works include acquiring larger amounts of data with more significant variations to expand evaluation of our approach, 
and investigating the effectiveness of our approach in downstream applications such as AI-augmented robotic surgery education. 


\subsubsection{Acknowledgement.}
This project was supported by CUHK Shun Hing Institute of Advanced Engineering (project MMT-p5-20), CUHK T Stone Robotics Institute, Hong Kong RGC TRS Project No.T42-409/18-R, and Multi-Scale Medical Robotics Center InnoHK under grant 8312051.



\bibliographystyle{splncs04}
\bibliography{ref}

\end{document}